\documentclass[11pt,a4paper]{article}
\usepackage[hyperref]{acl2021}
\usepackage{times}
\usepackage{latexsym}
\usepackage{graphics}
\usepackage{threeparttable}
\usepackage{graphicx}

\usepackage{microtype}

\aclfinalcopy % Uncomment this line for the final submission
\title{UTNLP at SemEval-2021 Task 5: A Comparative Analysis of Toxic Span Detection using Attention-based, Named Entity Recognition, and Ensemble Models}
\author{
  Alireza Salemi, Nazanin Sabri, Emad Kebriaei, Behnam Bahrak, Azadeh Shakery \\
  School of Electrical and Computer Engineering, College of Engineering\\ University of Tehran, Tehran, Iran\\
  \small{\texttt{\{alireza.salemi,nazanin.sabri,emad.kebriaei,bahrak,shakery\}@ut.ac.ir}}
}
\date{}

\begin{document}
\maketitle
\begin{abstract}
Detecting which parts of a sentence contribute to that sentence's toxicity---rather than providing a sentence-level verdict of hatefulness---would increase the interpretability of models and allow human moderators to better understand the outputs of the system. This paper presents our team's, \textit{UTNLP}, methodology and results in the SemEval-2021 shared task 5 on toxic spans detection. We test multiple models and contextual embeddings and report the best setting out of all. The experiments start with keyword-based models and are followed by attention-based, named entity-based, transformers-based, and ensemble models. Our best approach, an ensemble model, achieves an F1 of 0.684 in the competition's evaluation phase. 
\end{abstract}
% Being able to detect which parts of a sentence contribute to the toxicity of that sentence, rather than providing a sentence-level verdict of hatefulness, would increase interpretability of models, and allow human moderators to better understand the outputs of the system. In this paper, the methodology and the results obtained by our team, entitled \textit{UTNLP}, on the SemEval-2021 shared task 5 on toxic spans detection are presented. We test multiple models and word embeddings and report the best setting out of all models. The experimentation starts with keyword-based models, and is followed up by attention-based, named entity-based, transformers-based, and ensemble models. Our best model, an ensemble model which works based on intersecting the results of different models, achieves an F1 of 0.684 in the evaluation phase of the competition. 
 
\section{Introduction}
\label{sec:introduction}
When social media platforms were first introduced, they allowed users to post content on any topic they wished, without restricting the type of content they were allowed to put out. This absence of restrictions, along with the anonymity of users through these platforms \cite{pinsonneault1997anonymity, mondal2017measurement}, resulted in the spread of offensive language and hate speech online. While one might think there are only a small number of users who produce these types of hateful content, it has been shown that if social media platforms are left unmoderated, over time, the language of the community as a whole will change such that it highly correlates with the speech of hateful users rather than non-hateful ones \cite{mathew2020hate}. Given the huge number of social media postings every day, manual moderation of these platforms is not a possibility. As a result, many researchers began to study automatic hate speech detection.\\
Most studies on hate speech detection only provide labels at the sentence level, showing whether the construct as a whole is toxic or not. But these types of models, offer little explanation as to why the class was predicted, making it hard for human moderators to interpret the results \cite{semeval_2020_toxic_span}.\\ 
In an attempt to solve the aforementioned issue, we took part in SemEval-2021 shared task 5 \cite{pav2020semeval}, where we aim to detect which spans of a sentence cause it to become toxic. Our contributions are as follows: We begin our experimentation by evaluating a random baseline. Next, we test keyword-based methods, trying to find if toxic spans often include words that are known as hateful or negative in available word lists. We then test attention-based models, building on the hypothesis that what the attention model learns to focus on when detecting toxic speech, are the toxic spans. Afterwards, we look at the issue as a named entity recognition problem, by considering \emph{toxic} as a named entity category. Finally, we fine tune \textit{T5-base} and explore the possibility of looking at the task as a text-to-text problem. We compare different neural network architectures and embeddings, and report the model with the best performance. Additionally, we experiment with some hand-crafted features and evaluate their effectiveness in detecting toxic spans. Our best result, an ensemble of named-entity-based models, achieves an F1 of 0.684.

\section{Related Work}
\label{sec:related_work}
In this section we provide a brief overview of studies on hate and toxic speech detection, followed by work on span detection in different sub-fields. 
\subsection{Hate Speech}
Hate speech is defined as ``any communication that disparages a person or a group on the basis of some characteristic such as race, color, ethnicity, gender, sexual orientation, nationality, religion, or other characteristics" \cite{nockleby2000hate}. However, no clear distinction between toxic and hateful speech has been provided in the scientific literature \cite{d2019towards}. There are quite a few surveys on the topic of hate speech detection. \cite{schmidt2017survey}, describes available methods, features, and models for such a task. Another survey conducted in 2018 \cite{fortuna2018survey}, offers another view of the current state of the field, as well as suggesting ways in which hate speech detection could advance further. Other surveys on the topic published in 2020 include: \cite{naseem2020survey} which examines the impact of pre-processing on the performance of hate speech models. Corpora and resources for the task are studied in \cite{poletto2020resources}. Additionally, throughout the years, many shared tasks have been organized to help propel studies in the field \cite{vu2020hsd, bosco2018overview, basile2019semeval}. In addition to the classification of hate speech, significant effort has been put into the analysis of the target of hate \cite{silva2016analyzing, elsherief2018hate}.\\
Although numerous models have been tested, \cite{grondahl2018all} argues that when it comes to hate speech detection, the model is less important than the labeling criteria and the type of data. In confirmation of the importance of labeling, \cite{arango2019hate} also finds that models trained on annotations done by experts outperform systems trained on amateur annotations.
\subsection{Span Detection}
Named entity recognition (NER), code-switching detection, quotation detection, and key-phrase extraction are among many tasks that involve span identification.\\
\cite{chen2020improving} employs SpanBERT \cite{joshi2020spanbert} accompanied by a sequence tagging model to detect erroneous spans, proceeding to use detected spans to perform error correction. The combination of conditional random fields (CRF) \cite{lafferty2001conditional} and attention mechanisms \cite{vaswani2017attention} are explored in \cite{xu2020aspect} to explore aspect sentiment classification. The study finds that the use of multiple CRFs (to some limit) does improve performance. In \cite{papay2020dissecting} the authors look into systems to predict the performance of span identification tasks. To do so, BIO labels are used, and it is found that BERT \cite{devlin2018bert} helps when there is little data, while CRF is of great help in hard cases. In addition, the frequency of spans is found to help while length hurts the performance of the model. Furthermore, LSTMs \cite{hochreiter1997long} are reported to require large amounts of data to learn. \cite{tang2019progress} explores using fine-tuned BERT with attention models to extract keywords, showing how such models could enable the text classification model to be human interpretable. 

\section{Data}
\label{sec:data}
In this section, we will provide a brief description of the datasets utilized in this study. We will begin with our main dataset in which span-level toxicity has been labeled (\ref{sec:data:main}), next we look at other datasets that were used to better train our models, namely the hate word list that was used (\ref{sec:data:wordlist}) and the sentence-level hate speech data (\ref{sec:data:hatespeech}). 

\subsection{Main Task Dataset: Toxic Spans}
\label{sec:data:main}

The main dataset used in this study is that of the SemEval 2021, Toxic Span detection task \cite{semeval_2020_toxic_span, borkan2019nuanced}. In this dataset, which was built upon \textit{Civil Comments} \cite{DBLP:journals/corr/abs-1903-04561}, toxic word sequences (for sentences in the English language) have been labeled. In other words, labels are indexes of characters that are toxic in each sentence. There are a total of 8,629 sentences in the dataset, 8,101 of which include at least one annotated toxic span, and the rest have none. Sentences are on average 35 words long. The word-length of the toxic spans varies from one to 176 words. Toxic spans are, on average, 2.29 words long.\\ %Figure \ref{fig:length_of_spans} displays the distribution of the span lengths, limiting the length to 10, as only 2\% of the annotated toxic spans are longer than 10 words. \\
% \begin{figure}
% \centering
% \includegraphics[scale=0.4]{Figures/Figure1.png}
% \caption{Word-length of toxic spans in the dataset}
% \label{fig:length_of_spans}
% \end{figure}
There are some issues with regard to the quality of the annotations in the dataset. Table \ref{table:toxic_span_example} shows some examples of annotated comments in the dataset. While the first sentence is satisfactorily annotated, the second and third examples display issues with the labels in the dataset. More concretely, in the second example we can see that the indexes result in poorly separated and broken up words. Additionally, the anotated words are not toxic. In the third example we see that some of the words which do have a toxic connotation are not included in the annotation. While these examples are not extremely common in the dataset, these types of issues make automatic detection of such spans much more difficult. %they are certainly enough to be noticeable. These types of issues make automatic detection of such spans much more difficult. 

\begin{table*}[hbt!]
\begin{center}
  \caption{Examples of comments and the annotated toxic spans in the dataset}
  \label{table:toxic_span_example}
  \begin{tabular}{c c}
    \hline
    Text& Toxic Spans\\
    \hline
    what load you trump chumps just do not have any idea how to\\ deal with reality you have terrible judgment and pick exceptionally\\ idiotic arrogant leaders trump admitted he fired comey to stop \\the russia investigation man is he stupid. & ['idiotic', 'man is he stupid'] \\\hline
    except for one thing they are liars they only care about being thugs & ['r one th']\\\hline
    what harm have you ever heard of someone getting attacked\\ by bear while taking dump in the woods please does just owning gun make\\ someone paranoid and pu55y at the same time & ['harm']\\\hline
\end{tabular}
\end{center}
\end{table*}

\subsection{Datasets Used for Training}

To better train our models, we made use of several auxiliary datasets. 

\subsubsection{Word-list Dataset}
\label{sec:data:wordlist}

One of the methods tested in this study is based on word-matching. In other words, we check whether each word in the sentence is among hateful words and if so predict its label to be toxic. While this method is rather simple and we acknowledge that not all hate words are toxic and they could simply be used as a joke, we consider this method as a good first step to help us better understand the task at hand. As a result we need to use a list of hate words. For that purpose, we used a list of 1,616 unique hate words found on Kaggle \cite{hate_word_list}.  

\subsubsection{Hate Speech Dataset}
\label{sec:data:hatespeech}

To be able to train our attention-based models (\ref{sec:methods:attention}) we needed to have sentence-level annotated data. Thus we used the \textit{Civil Comments} dataset \cite{Jigsaw_Unintended_Bias_in_Toxicity_Classification}. The fact that this dataset and our main dataset have the same domain is the reason why this specific dataset was selected. In this dataset, each sentence is labeled with a number between 0 and 1, representing how hateful the text is. We consider sentences with scores above 0.5 to be hateful, and consider the rest as non-hateful. We then create a balanced sample of 289,298 sentences to train our model. The average length of sentences in this dataset is 48.12 words which is slightly longer than the sentences in the main dataset (\ref{sec:data:main}).

\section{Methodology}
\label{sec:methods}

In this study we have tested and compared various models to perform toxic span detection. In this section we will go over the structure and hyperparameters of these models. The codes of all models are publicly available on GitHub\footnote{\url{https://github.com/alirezasalemi7/SemEval2021-Toxic-Spans-Detection}}. 

\subsection{Attention-based Methods}
\label{sec:methods:attention}

We begin with the intuition that if a model with an attention layer is trained to detect hate speech at the sentence-level, the words the attention layer would learn to place importance on, would be the hateful words and spans. Consequently, we create a model made up of the following three layers:\\
\textbf{(1)} BERT-Base (Uncased) Layer which encodes our input texts. \\
\textbf{(2)} Attention Layer which is meant to be used for the aforementioned purposes\\
\textbf{(3)} Dense Layers which connect the attention outputs to two output nodes, detecting if the text is hateful or not.\\
% After the model has been trained, we use outputs of the attention layer (details of the model have been explained in Appendix \ref{appendix:1}). 
To train this model we have two training stages:
\begin{itemize}
    \item Sentence-level classification of hate speech
    \item Span-level detection of toxic spans
\end{itemize}

First we perform pre-processing by removing all punctuations (except those in the middle of words such as a\$\$hole), and lower-casing all words.\\
Next, the aforementioned model is designed. The BERT-Base (Uncased) layer has an input size of 400 tokens (clipping the input at 400 tokens and dropping the rest). The outputs of this layer are embedding vectors with a hidden size of 768 corresponding to the 400 input tokens. The second layer is an attention layer (attention matrix size = 4096) with a Relu activation function. Our last layers are two fully connected layers (4096 nodes) with dropout of 0.1. There are two neurons in the final layer, the objective of which is to detect whether the sentence is an instance of hate speech or not. The model is trained for 10 epochs with the Adam optimizer and a learning rate of 0.001. We freeze the weights of the BERT layer during this training process as we find through experimentation that fine-tuning BERT in this stage results in lower performance of our model in the toxic span detection task.\\
%  do not fine-tune BERT during this training process as we find through experimentation that fine-tuning BERT in this stage results in lower performance of our model in the toxic span detection task.\\
Once the model has been trained, we input our sentence and if our sentence-level detector predicts the sentence to be non-hateful we move on and produce a blank output as our toxic span. If, however, the model detects the sentence to be hateful, we extract the attention values and calculate the attention score of each word. If a word is made up of multiple subwords, we average the values of all subwords. After the attention scores have been calculated we use rule-based and machine learning models to label spans as toxic. These models are explained in Table \ref{table:attention_model_acc}. We begin by rule based models, selecting a percentage of spans with attention scores above a certain threshold (shown in Figure \ref{fig:percentage_and_thresh}). Additionally, we test different machine learning models with various sets of features. Our results are shown in Section \ref{subsec:results:attention}.

\subsection{Named Entity-based Methods}
\label{sec:methods:NER}

\begin{figure*}
\centering
\includegraphics[scale=0.55]{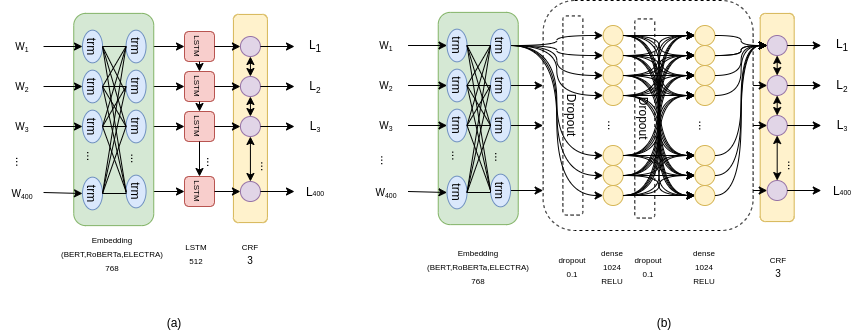}
\caption{The architecture of the named entity-based models. (a) displays a version of the model in which the two dense layers in (b) have been replaced by an LSTM layer. The results of both versions are shown in Table \ref{table:ner_model_acc}.}
\label{fig:ner_model_arch}
\end{figure*}

Our second intuition is to look at this problem as one similar to NER. As such, our toxic span label can be looked at as another NER label. We considered toxic, non-toxit and padding as labels and applied CRF to this NER task. The padding label was added to reduce the model bias toward the non-toxic class. \\
Our model is depicted in Figure \ref{fig:ner_model_arch}. We train the model for 2 epochs with a learning rate of $3\times10^{-5}$. In contrast to the previous method, the embedding layer is fine-tuned during our training process. Our tests on these models are shown in Section \ref{subsec:results:ner}. 

% \subsection{Google's T5}
% \label{sec:methods:t5}

% T5 is a text-to-text model, trained on a number of natural language understanding tasks \cite{raffel2019exploring}. We attempt to see if this problem can be modeled as a text-to-text problem, matching sentences to their toxic spans \cite{site:t5code}. 

\subsection{Ensemble Models}
\label{subsec:methods:ensemble}
Finally, we test two methods of combining the outputs of various models in order to achieve a better performance on the task. As previously mentioned, the expected outputs of the task are numerical indexes of the parts of the string which are believed to be toxic. Consequently, the first method of mixing could be voting, where if the majority of the models vote for one index, the index is included in the final selection. The second method is based on calculating the intersection of outputted indexes of all three models. In other words, only adding an index if it is detected by all three models. The results are shown in Section \ref{subsec:results:ensemble}. 

\section{Results}
\label{sec:results}
In this section we will report the results of the models introduced in the previous section (\ref{sec:methods}) on the toxic span detection task. Per the competition evaluation instructions, for all models the F1 score is reported. 

\subsection{Random Baseline}

To help us better understand the complexity of the task at hand, we start with a random baseline. In this method, we first split each sentence into words (using NLTK's functions) and then randomly label each word as toxic or not. We observe that this baseline F1-score for the task is 0.17. %The results are shown in Table \ref{table:random_results}.

% \begin{center}
% \begin{table}
%  \centering
%  \caption{Random baseline results}
%  \label{table:random_results}
%  \begin{tabular}{||c c||} 
%  \hline
%  Tokenization Method & F1\\ [0.5ex] 
%  \hline\hline
%  Space & 0.173 \\ 
%  \hline
%  NLTK & 0.179\\
%  \hline
% \end{tabular}
% \end{table}
% \end{center}

\subsection{Keyword-based}

The second simple method we test is a word-matching one. Our intuition is that toxic spans will likely include hateful or negative words. Thus we begin with a list of hate words and label any word found on the list as toxic and label the rest as nontoxic. This method results in an F1-score of 0.332 which is almost twice that of the random baseline, showing that while not all hate words are toxic and not all toxic spans are hate words, there is still a considerable amount of overlap. We further test if most words in toxic spans will have a negative sentiment value. Thus we repeat the same method, this time labeling anything with a negative sentiment as toxic. To detect the sentiment score of each word we use \textit{TextBlob} \cite{loria2018textblob}. We see that this method achieves an F1 of 0.378, outperforming the aforementioned technique. Finally we mix the two methods (labeling both hate words and words with negative sentiment as toxic), and achieve an F1-score of 0.418. 

% \begin{center}
% \begin{table}
%  \centering
%  \caption{Results of the word-based methods}
%  \begin{tabular}{||c c c||} 
%  \hline
%  Word Groups & Tokenization & F1\\ [0.5ex] 
%  \hline\hline
%  Hate & NLTK &  0.332\\ 
%  \hline
%  Negative Sentiment & NLTK & 0.378 \\ 
%  \hline
%  Both & NLTK & 0.418 \\ [1ex] 
%  \hline
% \end{tabular}
% \label{table:keyword_based_results}
% \end{table}
% \end{center}

% \subsection{Machine Learning Models}

\subsection{Attention-based}
\label{subsec:results:attention}

\begin{figure*}
\centering
\includegraphics[scale=0.26]{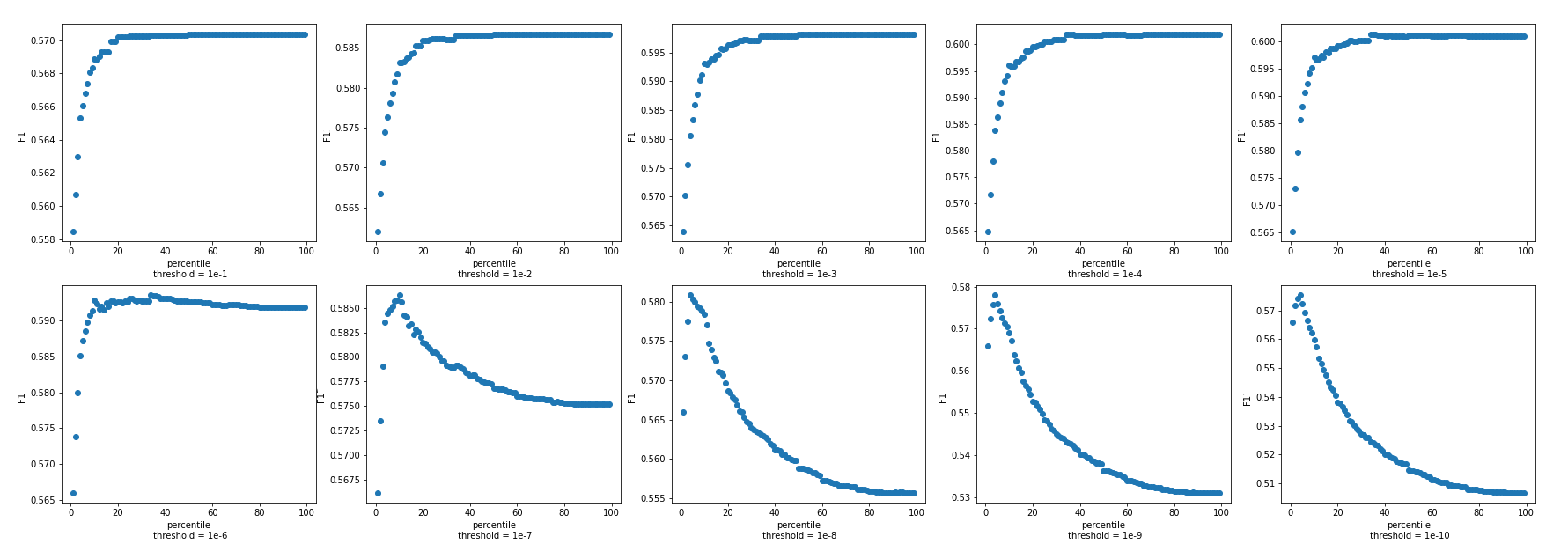}
\caption{The effects of placing various thresholds on the minimum value of attention scores allowed to be selected and the percentage of top scores that have been selected on the F1-score of the toxic span detection task. Each plot displays one threshold value, and the x-axis in each plot is the percentile of scores we select and the y-axis is the F1 value achieved by this combination of threshold and percentile.}
\label{fig:percentage_and_thresh}
\end{figure*}

As mentioned in Section \ref{sec:methods}, the intuition behind the attention-based model is that the model which learns to detect hate speech, would learn to pay more attention to the hateful spans in the text. Consequently, we test this idea in Table \ref{table:attention_model_acc}. We can see that the rule-based attention selection method outperforms other span selection techniques. To select the best set of rules for the model, we test both the percentage of top-words (with respect to attention) which we consider for selection, and the threshold we place on the minimum value of attention which is considered. As shown in Figure \ref{fig:percentage_and_thresh}, we can see that the top 75\% of attention scores with a threshold of $10^{-4}$ is the best set of hyper-parameters for the task.  

\begin{table*}[hbt!]
\begin{center}
\begin{threeparttable}
  \caption{Results of the attention-based models, the model structure is \textit{BERT + Attention + Dense} and we have tested out different span selection rules}
  \label{table:attention_model_acc}
\begin{tabular}{|c|c|c|c|c|c|c|c|}
 \hline
 \multicolumn{1}{|c|}{Model} & \multicolumn{4}{|c|}{Hate Speech Detection} & \multicolumn{2}{|c|}{Toxic Span Detection} \\
 \hline
 Span Selection Rules & Accuracy & Precision & Recall & F1 & F1 & F1 (Competition Evaluation)\\
 \hline
 R1 \tnote{a} &  &  &  &  & 0.601 & 0.609\\\cline{1-1}\cline{6-7} 
 R2 \tnote{b} &  &  &  &  & 0.601 & -\\\cline{1-1}\cline{6-7} 
 R3 \tnote{c} & 0.85 & 0.85 & 0.85 & 0.85 & 0.496 & - \\\cline{1-1}\cline{6-7} %
 Decision Tree \tnote{d}  &  &  &  &  & 0.360 & -\\\cline{1-1}\cline{6-7} % 
 Neural Network \tnote{e} &  &  &  &  & 0.354 & -\\
 \hline
\end{tabular}
\begin{tablenotes}
      \small
      \item[a] \textbf{R1}: selecting words with top 75\% of attention scores with threshold $10^{-4}$ and then removing stop-words
      \item[b] \textbf{R2}: R1 + removing positive sentiment words among the top 75\%
      \item[c] \textbf{R3}: R2 + adding all hate words (using the hate word list) in the sentence regardless of attention scores
      \item[d] \textbf{Decision Tree}: the input features of the model are: 1-attention score of word, 2-part of speech of the word, 3-sentiment of word 4-whether the word is a hate word or not (0/1)
      \item[e] \textbf{Neural Network}: the features inputted to the model are 1-attention score of word, 2-part of speech of the word, 3-sentiment of word, 4-whether the word is a hate word or not (0/1) - categorical features (e.g. POS) are modeled as learnable embeddings. 
    \end{tablenotes}
\end{threeparttable}
\end{center}
\end{table*}

Upon analysis of the results of the attention-based model, we find that the model performs well on the detection of single word spans (detecting 78\% of single-word spans in the evaluation dataset) but does not detect multi-word spans well (only detecting 16\% of such spans completely). This is because the distribution of attention scores are observed to be such that there is a large focus on one word and other words receive little attention values.\\
We further set up another experiment where we assumed that the true sentence-level labels were given. The model then predicted the toxic spans given these true labels achieving an F1 of 0.808. This shows that if the sentence-level classifier performed better, our model would have been able to get higher performance. Thus, more focus should be placed on obtaining higher accuracy in the sentence-level classification task.
% However, due to the lower accuracy and method of inference of our model the obtained results are much lower.

\subsection{Named Entity-based}
\label{subsec:results:ner}

Table \ref{tab:ner_results}, displays the results of our named entity based models. We can see that LSTM layers do not improve performance, and among various embeddings, RoBERTa outperforms the others in our 5-fold cross validation testings. However, BERT achieves better results in the competition's evaluation phase. 

\begin{table*}[hbt!]
\begin{center}
  \caption{Results of the named-entity based models evaluated using 5-fold cross-validation}
  \label{table:ner_model_acc}
  \begin{tabular}{c c c c c}
    \hline
    Embedding & Layers & F1 (train) & F1 (test)& F1 (Competition Evaluation)\\
    \hline
    BERT & CRF & 0.702 & 0.648 & 0.67\\\hline
    RoBERTa & CRF & 0.682 & 0.652 & 0.66\\\hline
    Electra & CRF & 0.687 & 0.646 & 0.65\\\hline
    BERT & LSTM + CRF & 0.668 & 0.62 & -\\\hline
    RoBERTa & LSTM + CRF & 0.669 & 0.647 & -\\\hline
    Electra & LSTM + CRF & 0.678 & 0.641 & -\\\hline
\end{tabular}
\label{tab:ner_results}
\end{center}
\end{table*}

\subsection{Google's T5}
\label{subsec:results:t5}

Another model we test is Google's T5 \cite{raffel2019exploring}. To test the T5 model, we use hugging-face’s \textit{T5-base} model\footnote{We were not able to test a larger version of the model due to system  constraints} and frame our problem as one where the context is the Tweet text and the answer is the text of the toxic spans to be detected. Our model achieves an F1 of 0.635 in the evaluation phase of the competition.    

\subsection{Ensemble Models}
\label{subsec:results:ensemble}

As described in section \ref{subsec:methods:ensemble}, we tested intersecting and using a voting scheme for the model outputs. More precisely, we perform these methods on the outputs of the following named entity based models:
\begin{itemize}
    \item[(1)] BERT + CRF
    \item[(2)] Electra + CRF
    \item[(3)] RoBERTa + CRF
\end{itemize}
We find that the competition evaluation F1 reaches 0.681 when we use voting of indexes, and 0.684 when the indexes are intersected. As can be seen both methods outperform all individual models. 

\section{Conclusion}
\label{sec:conclusion}
In this study we presented and compared various methods for toxic span detection. We examined the problem from various points of views reporting our results using each model. Our best system, an ensemble model, achieved an F1 of 0.684 in the SemEval-2021 Task 5 evaluation phase. Among the named-entity-based models, BERT+CRF performs best achieving an F1 of 0.67. Our attention-based model achieved an F1 of 0.609 in the competition's evaluation phase. Future work could focus on the improvement of the sentence-level detection in our attention scheme, as we showed improvement in that regard would improve this task's performance.

% \pagebreak

\bibliographystyle{acl_natbib}
\bibliography{anthology,acl2021}

% \appendix
% \section{Explaining How We Train the Attention-based Model}
% \label{appendix:1}

% To train this model we have two training stages:
% \begin{itemize}
%     \item Sentence-level classification of hate speech
%     \item Span-level detection of toxic spans
% \end{itemize}

% First we perform pre-processing by removing all punctuations, and lower-casing all words. Next we create a model with three layers. The first layer is a BERT-Base (Uncased) layer with an input size of 400 tokens (clipping the input at 400 tokens and dropping the rest). The outputs of this layer are embedding vectors with a hidden size of 768 corresponding to the 400 input tokens. The second layer is an attention layer with a Relu activation function. Our last layer is a dense layer with dropout. There are two neurons in the final layer and the model is trained to detect whether the sentence is an instance of hate speech or not.\\
% Once the model has been trained, we then input our sentence and if our sentence-level detector, predicts the sentences as non-hateful we move on and produce a blank output as our toxic spans. If, however, the model detects the sentence to be hateful, we extract the attention values and calculate the attention score of each word. If a word is made up of multiple subwords, we average the values of all subwords. After the attention scores have been calculated we user rule-based and machine learning models to label spans as toxic. These models where explained in Table \ref{table:attention_model_acc}.

\end{document}